\documentclass{article}

\usepackage[final]{neurips_2026_vericode_competition}

\usepackage[utf8]{inputenc} 
\usepackage[T1]{fontenc}    
\usepackage{hyperref}       
\usepackage{url}            
\usepackage{booktabs}       
\usepackage{amsfonts}       
\usepackage{amsmath}        
\usepackage{amssymb}        
\usepackage{nicefrac}       
\usepackage{microtype}      
\usepackage{xcolor}         
\usepackage{graphicx}       
\usepackage{fancyvrb}       
\usepackage{pgfplots}       
\usepackage{float}          
\pgfplotsset{compat=1.18}
\DeclareUnicodeCharacter{2192}{\ensuremath{\rightarrow}}
\DeclareUnicodeCharacter{2190}{\ensuremath{\leftarrow}}
\DeclareUnicodeCharacter{2229}{\ensuremath{\cap}}
\DeclareUnicodeCharacter{2260}{\ensuremath{\neq}}
\DeclareUnicodeCharacter{2200}{\ensuremath{\forall}}
\DeclareUnicodeCharacter{2208}{\ensuremath{\in}}
\DeclareUnicodeCharacter{2209}{\ensuremath{\notin}}
\DeclareUnicodeCharacter{2203}{\ensuremath{\exists}}
\DeclareUnicodeCharacter{2227}{\ensuremath{\wedge}}
\DeclareUnicodeCharacter{00AC}{\ensuremath{\neg}}
\DeclareUnicodeCharacter{2205}{\ensuremath{\emptyset}}
\DeclareUnicodeCharacter{2218}{\ensuremath{\circ}}

\newcommand{\code}[1]{\texttt{#1}}
\newcommand{\Lzero}{L_0}
\newcommand{\Hzero}{H_0}

\title{Par for the Proof:\\
Agentic Compression of Lean~4 Proofs under Exact Compiler Feedback}

\author{%
  Supreeth Rao \quad Anirudh Gupta\thanks{Student at the Indian Institute of Science (\texttt{anirudhgupta@iisc.ac.in}); this work was done at Emergence AI.} \quad Ajay Kumar Nair \quad Manan Tayal \\
  Emergence AI \\
  \texttt{\{srao, anirudh.gupta, ajay.nair, mtayal\}@emergence.ai}
}

\begin{document}

\maketitle

\begin{abstract}
  We describe our closed-track entry to the Lean Refactor Arena, where Lean~4 proofs must be rewritten to be shorter (fewer tokens) and cheaper to check (fewer heartbeats) while still compiling on every listed toolchain, for at most US\$3 of API spend per problem. We treat the task as a search against an exact, compiler-defined cost. A frontier model (gpt-6.1-sol) proposes proofs. A Lean metaprogram inserted into the target file measures each candidate as the competition does, at the cost of elaborating one declaration, and the best candidate so far is replaced only by one with a strictly higher three-axis score. A deterministic Lean optimizer then polishes the winner with verified local rewrites, and a reproduction of the competition scorer judges it. On the 60-problem final benchmark, our submission scores a combined \textbf{87.88} on the arena (length $-80.2\%$, heartbeats $-83.4\%$, all 67 cross-toolchain checks passed), at \$2.83 per problem on average and \textbf{\$2.97} at most. The score tracked the number of verified evaluations the budget bought: a cheaper model per turn, a session policy that spends the whole allowance, and half-price Flex tokens each raised it, while more information, more structure, or forced diversity did not.
\end{abstract}

\section{Introduction}
\label{sec:intro}

Proofs written for correctness, increasingly by agents, are often long and slow to check, which makes a library slower to build and harder to maintain across Lean releases. The Lean Refactor Arena~\cite{lra2026} makes proof economy measurable. Each problem is a theorem with its original proof, taken from a real Lean~4~\cite{demoura2021lean4} repository (Strata, PhysLib, CSLib, ArkLib, LeanPool) or from PutnamBench~\cite{tsoukalas2024putnambench}. A submission replaces the proof and is scored on three things: how much shorter it is, how much cheaper it is to elaborate, and how many of the problem's other Lean toolchains it still compiles on.

Three properties of the task shape our approach. Length is counted by Lean's own notation-aware lexer, so the true count exists only inside Lean with the file's imports loaded. The heartbeat axis is unbounded below: a two-token \code{by grind} that triples elaboration cost scores worse than the original, so every candidate must be measured, not estimated. And the closed track caps API spend at \$3 per problem, so the question is how far a fixed budget can move a proof.

We therefore treat proof compression as a search against an exact, compiler-defined cost under a fixed budget. A frontier model proposes proofs and edits, a deterministic Lean optimizer adds fine-grained rewrites, and Lean measures each candidate with the competition's own definitions. A strict keep-best incumbent ranks the candidates on the exact three-axis score, and an independent judge re-scores the winner. Two engineering problems follow: making each measurement exact and cheap, and turning as much of the \$3 as possible into such measurements. Our contributions are:
\begin{itemize}
  \item \textbf{A budget-aware agent harness.} The agent edits only the proof body, evaluates exact patches of archived candidates, and runs under heartbeat caps at the break-even point. Budget control is passive, and sessions restart with a flushed context from the current best (\S\ref{sec:agent}).
  \item \textbf{An account of what moves the score under a fixed budget}, from 25 full runs on the warm-up set, four on the final benchmark, and targeted pilots. The score tracked how many verified evaluations the budget bought; information, structure and forced diversity did not help (\S\ref{sec:results}).
\end{itemize}

\paragraph{Related work.}
LLM theorem proving has produced whole-proof generators, guided by informal drafts~\cite{jiang2023dsp} or repaired against the compiler's errors~\cite{first2023baldur}; provers trained on the proof checker's pass/fail signal with reinforcement learning or expert iteration~\cite{xin2024deepseekprover15,ren2025deepseekprover2,lin2025goedel,wang2025kimina}; and agents that call the proof assistant in a loop~\cite{yang2023leandojo,thakur2024copra,ospanov2025apollo} to close a goal that is still open. All three aim at a proof that does not yet exist. Our subject is a proof that already compiles: a frozen frontier model proposes edits, and every edit is measured against an exact, continuous three-axis cost rather than a pass/fail signal, under a fixed spend cap. The closest work is ImProver~\cite{ahuja2025improver}, which rewrites Lean proofs for user-chosen metrics such as length and readability; we optimize compiler-measured length and heartbeats together with cross-version compatibility. Our deterministic optimizer is a superoptimization-style search with delta-debugging-style deletion, and its suffix collapse borrows from proof-search automation such as Aesop~\cite{limperg2023aesop}. The harness follows ReAct~\cite{yao2023react} and SWE-agent's case for a narrow agent--computer interface~\cite{yang2024sweagent}. The Lean side is built on Lean~4's metaprogramming framework~\cite{metaprogramming2024}, as is LeanAide~\cite{agrawal2022leanaide}.

\section{Task and objective}
\label{sec:task}

Take a problem with reference length $\Lzero$ (tokens) and reference heartbeats $\Hzero$ (thousands of heartbeats, kHB). A candidate of length $L$ and heartbeats $H$ that compiles on $c$ of the problem's $k$ other toolchains scores
\begin{equation*}
  S \;=\; \tfrac{1}{3}\Big[\,100\big(1 - \tfrac{L}{\Lzero}\big) \;+\; 100\big(1 - \tfrac{H}{\Hzero}\big) \;+\; 100\,\tfrac{c}{k}\,\Big],
\end{equation*}
where the compatibility term is 100 when $k=0$. The combined score is the mean of $S$ over \emph{all} benchmark problems. A problem that fails a gate or does not compile on its default toolchain scores 0 on every axis. The objective is thus to maximize $S$ on every problem, trading tokens against heartbeats while keeping every listed toolchain compiling.

One token is worth exactly $\Hzero/\Lzero$ heartbeats. This per-problem exchange rate ranges from 1.5 to 16{,}013\,kHB per token on the final benchmark (median 29), so the same edit can be a gain on one problem and a loss on another. The final benchmark has 60 problems, ten from each of six sources, on eight Lean toolchains (v4.26 to v4.35-rc3), and lists 67 compatibility toolchains beyond the defaults.

\section{Approach}
\label{sec:approach}

\subsection{Overview: a verified search under a budget}

\begin{figure}[t]
  \centering
  \includegraphics[width=0.66\linewidth]{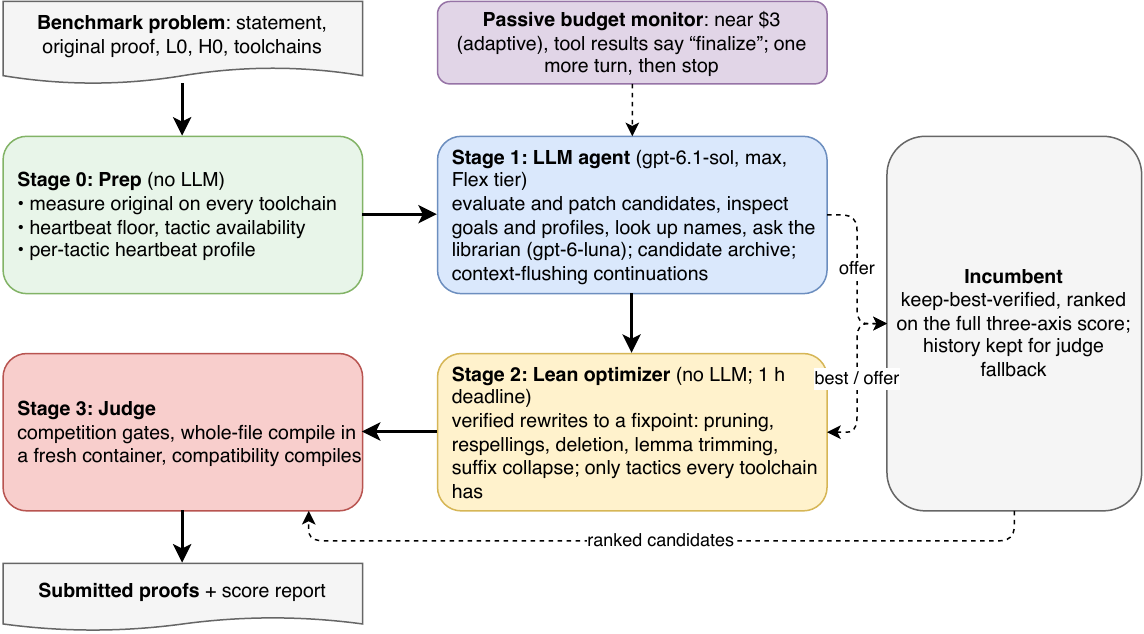}
  \caption{Per-problem pipeline. Stages 0--2 compile every candidate on a farm of warm, per-toolchain Lean processes (Figure~\ref{fig:infra}), and only Stage~1 spends API budget. Every candidate that reaches the submission has been verified by the compiler and re-scored by an independent reproduction of the competition scorer.}
  \label{fig:pipeline}
\end{figure}

Figure~\ref{fig:pipeline} shows how one problem is processed. Before any model call, the original proof is measured and profiled, and the facts the agent needs are assembled (Stage~0). The agent then proposes candidates, each of which is compiled and measured (Stage~1). A deterministic optimizer polishes the best candidate (Stage~2), and an independent judge re-scores it (Stage~3). Only the agent spends API budget.

All stages share one \emph{incumbent}. A candidate replaces the incumbent only if it compiles on the problem's default toolchain and its full three-axis score, compatibility included, strictly beats the current best. A poor candidate can never lower the score. The incumbent also keeps its history: if the judge rejects the best candidate, the next-best is judged, down to the original proof.

\subsection{Measuring the objective exactly and cheaply}
\label{sec:measure}

The competition publishes its leaderboard and submission-validation code but not its scoring worker. We reproduced the published parts directly: the scoring and its rounding, the comment stripper, the forbidden-pattern filter and the statement check. We reconstructed the measurement and checked it against the reference values.

\paragraph{Length} is a Lean lexer count, not a subword-tokenizer count (\code{tiktoken} matched 0 of 15 reference lengths). We run Lean's own token function inside the target file's environment, so notation declared by imports lexes correctly and comments cost nothing. Three corrections match the reference counts: a synthetic hole such as \code{?\_} counts as one token, a pipe (\code{<|}, \code{|>}) as two, and a \code{.} only when it stands alone as a focusing bullet.

\paragraph{Heartbeats.} Mathlib's \code{\#count\_heartbeats} is unavailable in Mathlib-free corpora. We instead bracket the declaration's elaboration with Lean's heartbeat counter, with asynchronous elaboration disabled. This counter works on every toolchain in the benchmark. The final benchmark ships no reference heartbeats, so our $\Hzero$ is our own measurement of each original proof.

\paragraph{One elaboration per candidate.} A whole-file compile in a fresh environment took about 4.5\,min in the case we timed, which would allow only a handful of evaluations per problem. Our measurement server is a Lean command spliced into the target file immediately before the theorem. Lean elaborates the file prefix (imports and everything above the theorem) once. The server then answers each request by elaborating a single declaration in that environment, without committing it. The median evaluation in the submitted run took 5.7\,s end to end (10th--90th percentile 2.9--13.3\,s over 13{,}126 evaluations). Heartbeats are read per thread, so up to four candidates are measured concurrently without interfering. A candidate counts only if it has no errors \emph{and} no \code{sorry}: \code{grind?} answers \code{grind => sorry} when it fails, which would otherwise look like the cheapest proof available. The same lexer and counter are used by the server, the optimizer and the judge, so every component reports the same numbers.

\paragraph{What else the server tells the agent.} A score alone does not say why a candidate failed or where its cost goes. When Lean checks a proof, it records, for every tactic, the goals before and after it and the proof term it produced (the \code{InfoTree}). The server reads this record to answer four kinds of question: what the open goals are at any point in any branch; which steps never touched a goal and can be deleted; which term a closer such as \code{simp} actually built, which often names the one lemma that could replace it; and how many tokens and heartbeats each step costs. Lemma search works the same way: names and goal shapes are looked up in the theorem's own environment, so every lemma the agent is offered exists at that point in the file and can be used as is. Because all of these answers come from Lean itself, nothing outside Lean ever parses or counts a proof. Appendix~\ref{app:lean} gives details.

\subsection{The agent}
\label{sec:agent}

The agent runs in our in-house agent runtime with all built-in tools disabled, and it writes only the proof body after \code{:=}. The harness adds the unchanged statement and maps error positions back to proof lines. Its tools (Appendix~\ref{app:tools}) fall into four groups:
\begin{itemize}
  \item \textbf{Evaluation:} compile and measure 1 to 4 proof bodies, or apply exact text replacements to an archived candidate. Both return the three axis scores, the projected score, per-toolchain compatibility, and located errors with the goal state.
  \item \textbf{Inspection:} goal states per tactic and branch, and per-tactic heartbeat and token profiles.
  \item \textbf{Retrieval:} name lookup, filtered name search, type-shape search, and a librarian, a smaller model (gpt-6-luna) with the same lookup tools that answers questions with verified names.
  \item \textbf{Memory:} an archive of every candidate by ID, and notes that persist across sessions.
\end{itemize}
The agent's first message holds the original proof with its measurements on every toolchain and its per-tactic heartbeat profile, the exchange rate $\Hzero/\Lzero$, a heartbeat floor (the statement proved by \code{sorry}), and the automation tactics available on every toolchain.

\paragraph{Ranking and early stopping.} Every candidate passes through the same gates, measurement, compatibility checks and incumbent. The agent sees each candidate's projected score, so it optimizes the true objective rather than raw length. Forbidden patterns and statement edits are rejected before compilation, with a one-line reason. Each candidate runs under a heartbeat cap at the break-even point, above which it cannot beat the incumbent, so failing searches stop early.

\paragraph{Budget control.} An early design that let the agent reason about its budget stopped at \$1.0--1.4 of \$3, after 2 to 8 compile attempts. The agent is now never told its budget. A passive monitor prices every response instead. Once spend plus $2.5\times$ the costliest turn so far would reach \$3, tool results tell the agent to evaluate once more and submit. One turn later, the session is stopped. The rule forecasts the next turns from past ones; it does not check each request against the remaining budget before sending it (\S\ref{sec:limits}).

\paragraph{Continuations flush context.} When the agent submits with budget to spare, or its context reaches 135k tokens, a new session starts from the current best. It receives the best proof, its profiles and goal states, and the research notes, but no attempt history: a digest of recent attempts anchored earlier sessions on micro-variants of the best. After three sessions in a row without a new best, the next session restarts from the original proof, shown the best's cost as the bar but not its text.

\paragraph{Knowledge and isolation.} The system prompt holds one skill document of about 170 lines. The document covers the scoring and exchange rate, a workflow, general compression patterns and the rules, and contains no benchmark proof or problem-specific hint. Nothing is carried between problems or runs: every search starts from the benchmark's reference proof. Infrastructure faults end a session rather than appearing to the model as failing proofs, and all search state is journaled, so an interrupted proof keeps its spend and its best.

\subsection{Deterministic polishing}
\label{sec:polish}

The optimizer is a deterministic search over verified edits of the agent's best proof, under a one-hour deadline. It minimizes $J = L/\Lzero + H/\Hzero$, compared exactly in integers. A move is accepted only if the result elaborates without \code{sorry}, keeps the statement unchanged, uses only tactics that parse on every listed toolchain, and lowers $J$ by more than a noise margin. Phases run cheapest first, each shrinking what later ones must search:
\begin{itemize}
  \item deleting steps that never met a goal;
  \item verified respellings, and merging identical branch arms;
  \item deleting steps or their dependency closures, delta-debugging style;
  \item trimming lemma lists to what \code{simp?} and \code{grind?} report;
  \item converting tactic blocks to terms;
  \item collapsing a block's last steps into one probe of a cost-ordered ladder of closers (\code{rfl}, \code{omega}, \code{simp}, \code{grind}, \ldots).
\end{itemize}
Every edit replaces a span located by Lean's own source positions, so the rest of the proof is left exactly as it was. Appendix~\ref{app:trace} traces the optimizer on a benchmark proof.

\subsection{Independent verification}

The recorded score comes from a separate judge that follows the competition's procedure. It applies the gates and splices the candidate into a fresh copy of the corpus file, locating the declaration by its text because the benchmark's line numbers are stale against the pinned commits. It then compiles the whole file in a fresh container and runs compile-only checks on the other toolchains.

\subsection{Infrastructure}
\label{sec:infra}

Each (corpus, toolchain) pair has a prebuilt container image. It holds the toolchain, the corpus at its pinned commit with the benchmark's modules compiled, Mathlib's prebuilt build artifacts and our Lean library, so no evaluation ever compiles Mathlib. On a Kubernetes cluster, each problem leases one warm process for its default toolchain and one per other listed toolchain. Every candidate that compiles on the default is checked on the others in parallel, so compatibility is part of every score the agent sees. The final benchmark used 127 such pods at a concurrency of 60 problems (Figure~\ref{fig:infra}, Appendix~\ref{app:infra}).

\section{Models}
\label{sec:models}

The submitted pipeline uses two models. Both are closed-weight models from OpenAI, and neither was fine-tuned:
\begin{itemize}
  \item \textbf{gpt-6.1-sol} drives the agent. It runs through OpenAI's Responses API at reasoning effort \emph{max}, on the Flex service tier.
  \item \textbf{gpt-6-luna} answers the agent's research questions as a librarian. It runs at reasoning effort \emph{medium}, on the standard tier, with at most 16 turns per question.
\end{itemize}
We set no temperature, top-$p$ or output-token limit, so provider defaults apply. The pipeline uses no embedding models, rerankers or LLM judges. Search, goal inspection and profiling are Lean metaprograms, and the judge is the compiler plus our reproduction of the competition scorer. During development we also tried gpt-6-astra, gpt-6-sol, GLM-5.2 and GLM-5.3; none contributed to the submitted proofs.

\section{Budget accounting}
\label{sec:budget}

\paragraph{Pricing.} Spend is computed at list prices (Table~\ref{tab:models}) from every response's reported usage: uncached input, cached input and output tokens, with reasoning counted as output. The organizers confirmed that published Flex and caching prices count as list prices. OpenAI bills nearly all uncached input as cache writes, so we price those tokens at the cache-write rate. The ledger sums every response in every session, including failed turns, discarded candidates and the librarian's calls. No other stage makes API calls. Compile-farm and image-build compute is not capped in the closed track and is not included.

\paragraph{Totals.} Table~\ref{tab:budget} summarizes the submitted run: \textbf{\$169.89} in total, \textbf{\$2.83} per problem on average, and a \textbf{maximum of \$2.97} (\code{RS\_jointAgreement\_of\_goodCoeffs\_card\_gt}); no proof exceeded \$3. Table~\ref{tab:perproblem} (Appendix~\ref{app:perproblem}) gives every problem's cost by model. Two proofs stopped well below the cap: \code{containsPhenoCompletions\ldots} (\$1.90) reached the 30-session limit on a four-token original that nothing beat, and \code{pauliContrDown\_mul\ldots} (\$1.91) lost its compile pod to preemption, which ended its search.

\begin{table}[t]
  \centering
  \footnotesize
  \begin{minipage}[t]{0.37\linewidth}
    \centering
    \caption{API spend of the submitted run at list prices (per problem: Table~\ref{tab:perproblem}).}
    \label{tab:budget}
    \setlength{\tabcolsep}{3pt}
    \begin{tabular}{lr}
      \toprule
      Problems & 60 \\
      gpt-6.1-sol (agent) & \$168.30 \\
      gpt-6-luna (librarian) & \$1.59 \\
      \textbf{Total} & \textbf{\$169.89} \\
      Mean / median & \$2.83 / \$2.87 \\
      Minimum & \$1.90 \\
      \textbf{Maximum} & \textbf{\$2.97} \\
      Over \$3 & 0 \\
      Billed $-$ metered & \$0.00 \\
      \bottomrule
    \end{tabular}
  \end{minipage}\hfill
  \begin{minipage}[t]{0.60\linewidth}
    \centering
    \caption{Model usage over the 60 problems. List prices are US\$ per million tokens (uncached input / cached input / cache write / output).}
    \label{tab:models}
    \setlength{\tabcolsep}{2.5pt}
    \begin{tabular}{lrr}
      \toprule
      & gpt-6.1-sol & gpt-6-luna \\
      \midrule
      Tier & Flex & standard \\
      List prices & 1/0.05/1.25/5 & 0.1/0.01/0.125/0.5 \\
      Responses & 10{,}679 & 2{,}368 \\
      Input tokens & 686.6M & 31.6M \\
      \quad cache reads & 643.9M & 25.6M \\
      \quad cache writes & 42.7M & 6.0M \\
      Output tokens (incl.\ reasoning) & 16.5M & 1.2M \\
      Cost: cache reads / writes & \$32.19 / \$53.33 & \$0.26 / \$0.75 \\
      Cost: output & \$82.75 & \$0.59 \\
      \bottomrule
    \end{tabular}
  \end{minipage}
\end{table}

\paragraph{Billing evidence.} Every call went through our self-hosted API gateway, which logs each call's billed cost. At the end of the run, every response in a proof's ledger is matched to a gateway call by its exact token counts. The pipeline emits a submission only if every proof's matched billed cost is at most \$3 and no billed cost under the run's dedicated key is left unattributed. In the submitted run all 60 proofs verified: the gateway billed \$169.889384 against \$169.889384 metered, with no unmatched responses, and every agent call was billed on the Flex tier.

\section{Reproduction}
\label{sec:repro}

The code archive submitted with the proofs contains the complete pipeline: the agent harness and prompt, the Lean library, the judge, and the build recipes for the container images. Its top-level README documents a single entry point that installs the pinned environment, builds the images and runs the pipeline. The Lean toolchains and corpus commits are pinned by the benchmark, and the Python dependencies by a lockfile. One configuration file fixes every setting of a run: models, efforts, service tier, prices, the \$3 cap, session limits and the optimizer deadline.

A rerun needs an API key used by no other client during the run, with about \$200 of credit, plus Docker and a Kubernetes cluster able to schedule about 130 pods at once. The submitted run took 4\,h\,19\,min at a concurrency of 60; expect 4--5 hours and about \$170 in total, at most \$3 per problem. Flex capacity refusals are retried, which can lengthen the wall-clock but not the cost.

\paragraph{What is not deterministic.} Sampling, provider-side nondeterminism, parallel tool-call ordering and retry timing all vary. We ran the submitted configuration once on the full benchmark. On the warm-up set, three runs of the final agent configuration scored 92.36, 92.31 and 92.24, and over our development history a single run has been reproducible to about $\pm 0.5$ combined. Individual problems vary by several points; for example, \code{SKI.parallelReduction\_diamond} scored 78.49--82.38 across our four final-benchmark runs. We expect a rerun to land within about half a point of 87.99, but not to reproduce the submitted proofs line for line.

\section{Results and analysis}
\label{sec:results}

\subsection{The submitted run}

Table~\ref{tab:corpus} summarizes the submitted run by corpus. All 60 problems compile and all 67 compatibility checks pass. Total length falls from 85{,}453 to 11{,}356 tokens, and total heartbeats from 7.44M to 0.58M\,kHB. Our judge and the arena agree to 0.11 points (87.99 vs.\ 87.88), and exactly on 29 of 60 problems. The largest per-problem difference ($-2.78$) is a single token that the arena's lexer counts twice.

\paragraph{By corpus.} PutnamBench's long competition proofs compress the most (97.86), and PhysLib the least (74.14). Three PhysLib problems are 4- to 12-token \code{decide +kernel} proofs whose cost is kernel reduction. They score 33.33, 49.99 and 43.34, and without them the combined score is 90.40.

\paragraph{The search.} The agent made 10{,}679 turns in 563 sessions and evaluated 13{,}429 candidates. Of these, 55\% compiled, 38\% failed, 5\% were stopped at their break-even heartbeat cap, and 2\% repeated an earlier candidate and were answered from a cache. The optimizer improved 17 of the 55 problems it finished on, adding +0.07 combined.

\paragraph{A characteristic rewrite.} The original \code{putnam\_1964\_b2} is a 294-line, 1{,}754-token argument. The agent recognized the statement as an instance of Mathlib's characterization of maximal intersecting families, \code{Set.Intersecting.is\_max\_iff\_card\_eq}, and reduced the proof to the arithmetic needed to apply it: 27 tokens and 797\,kHB, down from 134{,}499\,kHB (Figure~\ref{fig:example}, Appendix~\ref{app:perproblem}). Recurring moves include inducting on a hypothesis instead of on data, functional induction in place of induction followed by unfolding, merging mirrored branches with \code{first | A | B}, and replacing a \code{simp} set by the one lemma that states the goal.

\begin{table}[t]
  \centering
  \footnotesize
  \caption{The submitted run by corpus (ten problems each; judge scores, except the last row). Every problem passes compatibility on every listed toolchain. Cost is API spend at list prices.}
  \label{tab:corpus}
  \setlength{\tabcolsep}{3.5pt}
  \begin{tabular}{llrrrrr}
    \toprule
    Corpus & Toolchains & $\sum \Lzero \to \sum L$ & Len.\% & HB\% & Score & Cost \$ \\
    \midrule
    Strata & 4.26--4.31 & 7066 $\to$ 1525 & 80.00 & 84.71 & 88.24 & 28.92 \\
    PhysLib & 4.34.0--4.34.1 & 6057 $\to$ 807 & 68.77 & 53.66 & 74.14 & 26.52 \\
    CSLib & 4.30--4.34.1 & 4076 $\to$ 1210 & 70.96 & 88.77 & 86.58 & 28.71 \\
    ArkLib & 4.30--4.34 & 23027 $\to$ 2818 & 87.80 & 93.41 & 93.74 & 28.80 \\
    LeanPool & 4.34--4.35-rc3 & 17903 $\to$ 4122 & 78.99 & 83.20 & 87.40 & 28.79 \\
    PutnamBench & 4.26--4.27 & 27324 $\to$ 874 & 96.38 & 97.21 & 97.86 & 28.15 \\
    \midrule
    All 60 (our judge) & & 85453 $\to$ 11356 & 80.48 & 83.49 & \textbf{87.99} & 169.89 \\
    All 60 (arena, official) & & & 80.24 & 83.39 & \textbf{87.88} & \\
    \bottomrule
  \end{tabular}
\end{table}

\begin{table}[t]
  \caption{Our four runs on the final benchmark, each row's change relative to the row above. All use gpt-6.1-sol (max) with the gpt-6-luna librarian at a \$3 cap. ``Total \$'' is billed spend.}
  \label{tab:final-ladder}
  \centering
  \footnotesize
  \setlength{\tabcolsep}{4pt}
  \begin{tabular}{llrrrr}
    \toprule
    Run & Change & Comb. & Len.\,\% & HB\,\% & Total \$ \\
    \midrule
    1 & warm-up configuration, standard tier, no optimizer & 84.70 & 75.09 & 79.01 & 163.94 \\
    2 & + deterministic optimizer; infrastructure fixes & 85.74 & 76.54 & 80.68 & 161.43 \\
    3 & + heaviest problem searched (previously lost to a profiling failure) & 86.96 & 78.85 & 82.02 & 169.45 \\
    \textbf{4} & \textbf{+ Flex tier (submitted)} & \textbf{87.99} & \textbf{80.48} & \textbf{83.49} & \textbf{169.89} \\
    \bottomrule
  \end{tabular}
\end{table}

\paragraph{From first run to submission.} Table~\ref{tab:final-ladder} lists our runs on the final benchmark. Two changes account for most of the 3.3-point gain. The first was searching \code{putnam\_2019\_a1}, whose 3.2M\,kHB original had exhausted its pod during profiling (33.33~$\to$~98.37). The second was the Flex tier.

\subsection{What moved the score: evaluations per dollar}

We developed on the 15 warm-up problems, 13 of which reappear in the final benchmark. Table~\ref{tab:ladder} (Appendix~\ref{app:dev}) lists the warm-up changes that moved the score by more than noise. The large steps share a mechanism: each one increased the number of verified evaluations the \$3 buys, or stopped budget from going unspent.

\paragraph{A cheaper model per turn.} Switching from gpt-6-astra to gpt-6.1-sol was the largest single step on the warm-up set (+2.0, to 90.57). Sol was not stronger per turn: astra vs.\ sol scored 78.5 vs.\ 84.7 at turn~5, and 88.4 vs.\ 89.4 at turn~16. But an astra turn costs about three times as much as a sol turn, and sol reached astra's \$3 score at about \$1.

\paragraph{Spending the whole allowance.} A flat \$2.50 finalize threshold left budget unused, and it overshot when turns were large. The adaptive rule of \S\ref{sec:agent} replaced it. Since then no proof has exceeded \$3 in any full run, and in the submitted run 58 of 60 proofs spent at least \$2.72.

\begin{figure}[t]
  \centering
  \begin{tikzpicture}
    \begin{axis}[
      width=0.70\linewidth, height=3.2cm,
      xlabel={API spend per problem (US\$, billed)}, ylabel={Mean best score},
      xmin=0, xmax=3.1, ymin=70, ymax=90,
      xtick={0.5,1,1.5,2,2.5,3}, ytick={70,75,80,85,90},
      legend pos=south east, legend style={font=\scriptsize, draw=none, fill=none},
      tick label style={font=\scriptsize}, label style={font=\small},
      grid=major, grid style={gray!20},
    ]
      \addplot[thick, color=black, mark=*, mark size=1.2pt] coordinates {
        (0.25,79.63) (0.5,83.33) (0.75,85.07) (1,85.86) (1.5,86.65) (2,87.19) (2.5,87.63) (3,87.92)};
      \addlegendentry{Flex tier (run 4, submitted)}
      \addplot[thick, dashed, color=gray, mark=square*, mark size=1.2pt] coordinates {
        (0.25,71.22) (0.5,80.12) (0.75,82.07) (1,83.60) (1.5,85.19) (2,85.88) (2.5,86.46) (3,86.84)};
      \addlegendentry{standard tier (run 3, same harness)}
    \end{axis}
  \end{tikzpicture}
  \caption{Mean incumbent score over the 60 final-benchmark problems against billed spend per problem, measured by the search before polishing. A proof that stopped early keeps its last best.}
  \label{fig:curve}
\end{figure}
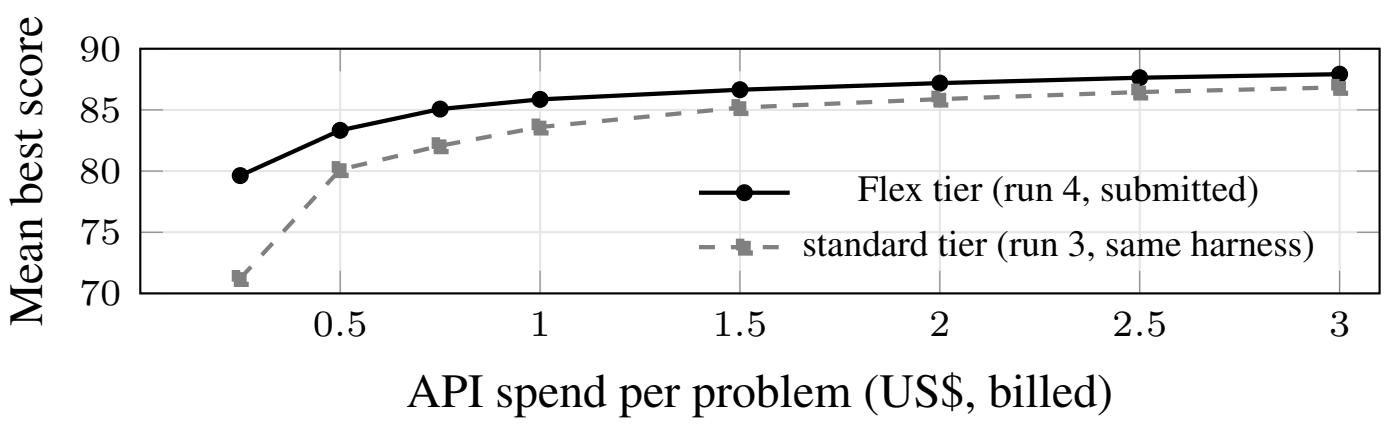

\paragraph{More tokens per dollar.} The Flex tier halves the price per token. On the final benchmark, runs 3 (standard tier) and 4 (Flex) used the same harness at the same billed spend (\$169.45 vs.\ \$169.89). Flex doubled the agent's turns (5{,}365 $\to$ 10{,}679) and evaluations (7{,}335 $\to$ 13{,}429) and raised the score by 1.03 (length $+1.63$, heartbeats $+1.47$), improving 49 of 60 problems. Figure~\ref{fig:curve} shows why: the score is concave in spend, and the Flex run reaches at \$1 (85.9) what the standard run reaches at \$1.5--2. Budget pilots agree: a \$6 cap on the six hardest problems scored 78.45 against 76.86 at \$3, and a \$9 cap on eight weak problems gained 1.33 per doubling. One long attempt also beats several short ones: two \$1.50 attempts, keeping the better, scored 86.42 against 88.27 for one \$3 attempt.

\subsection{What did not move the score}

Table~\ref{tab:ablations} (Appendix~\ref{app:dev}) lists the changes that did not raise the score beyond noise, or lowered it. More information did not help: lemma suggestions, offline pattern search, larger prompts and profiles on every new best all landed at noise. Nor did more structure: forced-diversity phases ($-1.2$), a planner--worker split ($-2.5$), parallel attempts ($-1.9$) and restarts all cost points. Lower reasoning effort, compiler-verified research ($-1.0$ to $-1.7$ on six hard final-benchmark problems) and a skill extended with 30 golf idioms (+0.14, within noise) did not move it either.

The trajectories explain why. On most problems the final proof's structure was fixed by turn 4--8, and 61\% of candidates were near-copies of the incumbent. Final proofs differ across runs (token similarity 0.42), yet they score within about a point of each other, and the best-of-runs oracle is only 0.4--0.5 above a single run. Extra information did not change which proofs the search reached, and forcing it towards different proof structures cost turns that refinement would otherwise have used. Knowledge was not the bottleneck either: better research tools cut lookup-only turns from 32--55\% to 5\% at no change in score. The golf idioms came from polishing our proofs with unlimited evaluations; their gains came from many measured micro-edits, which a \$3 agent cannot afford.

\subsection{The deterministic optimizer}
\label{sec:compressor-results}

Table~\ref{tab:optimizer} (Appendix~\ref{app:infra}) isolates the optimizer. Run alone on the warm-up originals, it lifts every problem above 33.33 (combined 43.68) but stops far short of the agent: local rewrites do not find restructurings such as Figure~\ref{fig:example}. Applied to the agent's output it adds +0.03 to +0.21, a gain that shrank as the harness improved. On the final benchmark it added +0.12 to run 1's proofs and +0.07 in the submitted run, mostly on long proofs. The agent's proofs already sit near the optimizer's fixpoint. We still run it because it costs no API budget and the incumbent prevents it from lowering a score. The Lean library's main contribution is the measurement behind every agent tool.

\section{Limitations}
\label{sec:limits}

\paragraph{Length.} Our remaining gap is on the length axis (80.5\% against 83.5\% for heartbeats). About a third of our final tokens are lemma names, and the rest is structure. Closing the gap needs structurally different proofs, which neither prompts, effort nor deterministic polishing produced.

\paragraph{Unseen theorems.} Most design decisions were made on the warm-up set, with a noise band of about $\pm 0.5$. Before the final runs we removed all warm-up-specific content from the prompt, at no cost on the warm-up set (92.36 vs.\ 92.31). Run~1 used the warm-up agent unchanged (prompt, skill, models and budget rule) on the standard tier, so it is a frozen evaluation: 84.70 overall, 82.99 on the 47 problems outside the warm-up set and 90.90 on the 13 inside it. Runs 2--4 left the prompt and skill unchanged and added the optimizer, fault handling, Flex and a higher session backstop (20 $\to$ 30); the submitted run scored 86.84 on the 47 and 92.16 on the 13. The gap mixes exposure during development with corpus difficulty (the three kernel-bound PhysLib problems are all new), and we cannot separate the two. The heartbeat axis uses our own $\Hzero$, since the final benchmark ships no reference heartbeats; the arena's heartbeat axis came out 0.10 lower than ours.

\paragraph{Budget guarantee.} The finalize rule is a forecast, not a worst-case bound: we set no output-token limit, so a single response could in principle cost more than the headroom left. In the submitted run the costliest response was \$0.078 (2.6\% of the cap), the smallest margin left under \$3 was \$0.03, and every proof's spend was confirmed against the gateway's billed cost. A hard guarantee would cap each request's output tokens at what the remaining budget can pay for.

\paragraph{Reproduction and faults.} A rerun needs the container images and a Kubernetes cluster (\S\ref{sec:repro}), and depends on provider availability as well as sampling. One proof's search ended at \$1.91 when its pod was preempted, and six agent sessions ended on provider stream errors and were continued from their best. The compile farm (127 pods for 4.3\,h) is not counted by the closed track.

\section{Conclusion}
\label{sec:conclusion}

An exact, cheap, compiler-side measurement lets a budget-capped frontier model do the conceptual work of proof compression on its own: propose a rewrite, check it against the compiler, and keep whatever strictly improves the score. That loop takes the Lean Refactor Arena's final benchmark to a mean 80.5\% shorter and 83.5\% cheaper to elaborate, with every proof compiling on every listed toolchain, at \$2.83 per problem on average and \$2.97 at most.

Across 25 warm-up runs and four final-benchmark runs, the score tracked the number of verified evaluations the budget bought, not the information or structure given to the agent. The gap that remains is on length, and closing it looks like it needs proofs built differently from the start, not more edits to the ones the search already finds.

\begin{ack}
This work was funded by Emergence AI, where all authors are employed. The authors declare no competing interests. No compute or API credits were provided by a model vendor. API usage was paid at list prices, and the compile farm ran on Emergence AI's cloud.
\end{ack}

\clearpage
{
\small
\bibliographystyle{plainnat}
\bibliography{references}
}

\clearpage
\appendix

\section{Per-problem results and spend}
\label{app:perproblem}

Table~\ref{tab:perproblem} lists every problem of the submitted run. All 60 problems score 100 on compatibility, so a problem's score is $(\text{Len.\%} + \text{HB\%} + 100)/3$. Every proof's cost was billing-verified against the gateway's per-call log.

\begin{table}[H]
  \caption{The submitted run (run 4), per problem; problems are listed by their last name component. $L$/$H$ are the judged length and heartbeats (kHB). ``S'' is the number of agent sessions and ``Evals'' the number of candidates evaluated. ``St.2'' marks problems where the deterministic optimizer improved the agent's best, and ``orig.'' where nothing beat the original. Cost is API spend at list prices per model, \code{gpt-6.1-sol} (Flex) and \code{gpt-6-luna} (standard), with retries, failed turns and discarded candidates included.}
  \label{tab:perproblem}
  \centering
  \resizebox{\linewidth}{!}{%
  \begin{tabular}{lrrrrrrrrrcrrr}
    \toprule
    Problem & $\Lzero$ & $L$ & $\Hzero$ & $H$ & Len.\% & HB\% & Score & S & Evals & St.2 & sol \$ & luna \$ & Total \$ \\
    \midrule
\addlinespace[2pt]\multicolumn{14}{l}{\emph{Strata (v4.26--4.31)}} \\
\code{InitsUpdatesComm} & 224 & 62 & 4180 & 177 & 72.32 & 95.77 & 89.36 & 19 & 191 &  & 2.79 & 0.035 & 2.82 \\
\code{denote\_suffix\_irrel} & 904 & 176 & 4183 & 441 & 80.53 & 89.46 & 90.00 & 8 & 179 &  & 2.88 & 0.017 & 2.90 \\
\code{denote\_Denotes} & 931 & 127 & 10925 & 1316 & 86.36 & 87.95 & 91.44 & 7 & 221 & \checkmark & 2.90 & 0.011 & 2.91 \\
\code{callOfLFunc\_eraseMetad\ldots} & 291 & 39 & 21294 & 1564 & 86.60 & 92.66 & 93.09 & 8 & 215 &  & 2.75 & 0.034 & 2.78 \\
\code{liftBVars\_typeCheck} & 209 & 22 & 17630 & 1677 & 89.47 & 90.49 & 93.32 & 12 & 331 &  & 2.90 & 0.017 & 2.92 \\
\code{liftBVars\_denote} & 972 & 291 & 8896 & 3030 & 70.06 & 65.94 & 78.67 & 7 & 159 & \checkmark & 2.91 & 0.018 & 2.92 \\
\code{typeCheck\_to\_HasTypeA} & 429 & 59 & 7096 & 1006 & 86.25 & 85.82 & 90.69 & 10 & 264 &  & 2.90 & 0.017 & 2.91 \\
\code{callOfLFunc\_constr\_dis\ldots} & 968 & 400 & 7353 & 1435 & 58.68 & 80.48 & 79.72 & 6 & 170 &  & 2.88 & 0.021 & 2.90 \\
\code{denote\_replaceMetadata} & 938 & 51 & 7274 & 571 & 94.56 & 92.15 & 95.57 & 10 & 167 &  & 2.88 & 0.010 & 2.89 \\
\code{denote\_ext} & 1200 & 298 & 4666 & 1569 & 75.17 & 66.37 & 80.51 & 7 & 170 &  & 2.94 & 0.006 & 2.95 \\
\addlinespace[2pt]\multicolumn{14}{l}{\emph{PhysLib (v4.34.0--4.34.1)}} \\
\code{containsPhenoCompletio\ldots} & 4 & 4 & 64051 & 64051 & 0.00 & 0.00 & 33.33 & 30 & 301 & orig. & 1.89 & 0.011 & 1.90 \\
\code{isPhenoClosedQ5\_viable\ldots} & 12 & 6 & 74041 & 74071 & 50.00 & -0.04 & 49.99 & 15 & 254 &  & 2.81 & 0.059 & 2.87 \\
\code{isPhenoClosedQ10\_viabl\ldots} & 12 & 6 & 79108 & 94924 & 50.00 & -19.99 & 43.34 & 26 & 241 &  & 2.84 & 0.038 & 2.87 \\
\code{pauliContrDown\_mul\_pau\ldots} & 181 & 50 & 212387 & 99647 & 72.38 & 53.08 & 75.15 & 7 & 153 &  & 1.87 & 0.040 & 1.91 \\
\code{pauliContr\_mul\_pauliCo\ldots} & 183 & 46 & 136539 & 71073 & 74.86 & 47.95 & 74.27 & 11 & 229 &  & 2.76 & 0.034 & 2.79 \\
\code{time\_deriv\_time\_deriv\_\ldots} & 1372 & 209 & 23939 & 3834 & 84.77 & 83.98 & 89.58 & 6 & 245 &  & 2.93 & 0.008 & 2.94 \\
\code{$\iota$\_timeOrderF\_superComm\ldots} & 1217 & 261 & 35580 & 4847 & 78.55 & 86.38 & 88.31 & 6 & 182 &  & 2.81 & 0.009 & 2.82 \\
\code{fundamental\_theorem\_of\ldots} & 851 & 93 & 10196 & 1144 & 89.07 & 88.78 & 92.62 & 14 & 152 &  & 2.73 & 0.065 & 2.79 \\
\code{inv\_convex} & 1025 & 64 & 211286 & 4020 & 93.76 & 98.10 & 97.29 & 10 & 167 & \checkmark & 2.75 & 0.023 & 2.77 \\
\code{cauchyBinet} & 1200 & 68 & 142163 & 2359 & 94.33 & 98.34 & 97.56 & 6 & 251 &  & 2.82 & 0.034 & 2.85 \\
\addlinespace[2pt]\multicolumn{14}{l}{\emph{CSLib (v4.30--4.34.1)}} \\
\code{progress} & 408 & 90 & 59947 & 6967 & 77.94 & 88.38 & 88.77 & 6 & 198 &  & 2.90 & 0.045 & 2.94 \\
\code{parallelReduction\_diam\ldots} & 407 & 118 & 1496 & 357 & 71.01 & 76.14 & 82.38 & 10 & 195 &  & 2.82 & 0.015 & 2.83 \\
\code{bisimilarity\_congr\_cho\ldots} & 251 & 47 & 3907 & 273 & 81.27 & 93.01 & 91.43 & 14 & 156 &  & 2.81 & 0.011 & 2.82 \\
\code{preservation} & 391 & 120 & 38121 & 7210 & 69.31 & 81.09 & 83.47 & 7 & 266 &  & 2.80 & 0.092 & 2.90 \\
\code{of\_consensus} & 433 & 210 & 11618 & 908 & 51.50 & 92.18 & 81.23 & 7 & 268 &  & 2.88 & 0.018 & 2.90 \\
\code{buchiFamily\_saturation} & 623 & 217 & 56919 & 4723 & 65.17 & 91.70 & 85.62 & 6 & 215 & \checkmark & 2.91 & 0.026 & 2.94 \\
\code{nonUniform\_step} & 368 & 55 & 39309 & 1287 & 85.05 & 96.73 & 93.93 & 10 & 175 &  & 2.84 & 0.022 & 2.87 \\
\code{flatten\_execution} & 312 & 113 & 27602 & 3782 & 63.78 & 86.30 & 83.36 & 7 & 281 & \checkmark & 2.85 & 0.021 & 2.87 \\
\code{subset\_inp} & 364 & 112 & 33186 & 1808 & 69.23 & 94.55 & 87.93 & 6 & 254 &  & 2.89 & 0.011 & 2.91 \\
\code{parallel\_diamond} & 519 & 128 & 8584 & 1063 & 75.34 & 87.62 & 87.65 & 6 & 151 & \checkmark & 2.71 & 0.031 & 2.74 \\
\addlinespace[2pt]\multicolumn{14}{l}{\emph{ArkLib (v4.30--4.34)}} \\
\code{interleaved\_affine\_gap\ldots} & 1353 & 93 & 7509 & 964 & 93.13 & 87.16 & 93.43 & 11 & 254 &  & 2.75 & 0.018 & 2.77 \\
\code{fold\_advances\_evaluati\ldots} & 1906 & 108 & 47452 & 4975 & 94.33 & 89.52 & 94.62 & 10 & 246 &  & 2.84 & 0.021 & 2.86 \\
\code{fiberwise\_dist\_lt\_imp\_\ldots} & 1463 & 163 & 16576 & 1427 & 88.86 & 91.39 & 93.42 & 7 & 267 & \checkmark & 2.92 & 0.024 & 2.94 \\
\code{master\_lemma} & 546 & 78 & 190178 & 2497 & 85.71 & 98.69 & 94.80 & 8 & 240 &  & 2.90 & 0.029 & 2.93 \\
\code{johnson\_bound\_alphabet\ldots} & 2440 & 581 & 84788 & 8269 & 76.19 & 90.25 & 88.81 & 6 & 243 & \checkmark & 2.90 & 0.023 & 2.92 \\
\code{dir\_close\_of\_many\_clos\ldots} & 5128 & 364 & 106549 & 3729 & 92.90 & 96.50 & 96.47 & 11 & 202 & \checkmark & 2.77 & 0.026 & 2.80 \\
\code{RS\_exists\_nonzero\_kern\ldots} & 2122 & 298 & 163963 & 3842 & 85.96 & 97.66 & 94.54 & 6 & 204 & \checkmark & 2.74 & 0.033 & 2.77 \\
\code{RS\_jointAgreement\_of\_g\ldots} & 5538 & 860 & 62530 & 4334 & 84.47 & 93.07 & 92.51 & 8 & 249 & \checkmark & 2.90 & 0.068 & 2.97 \\
\code{reduction\_perfectCompl\ldots} & 1426 & 58 & 44536 & 1791 & 95.93 & 95.98 & 97.30 & 7 & 242 &  & 2.88 & 0.026 & 2.90 \\
\code{ps\_coprime\_case\_constant} & 1105 & 215 & 38094 & 2340 & 80.54 & 93.86 & 91.47 & 6 & 268 & \checkmark & 2.90 & 0.022 & 2.93 \\
\addlinespace[2pt]\multicolumn{14}{l}{\emph{LeanPool (v4.34--4.35-rc3)}} \\
\code{erlv\_case12\_other\_d1\_b\ldots} & 2030 & 208 & 49666 & 775 & 89.75 & 98.44 & 96.06 & 6 & 173 &  & 2.74 & 0.046 & 2.78 \\
\code{matsumoto\_reduced\_aux} & 1699 & 338 & 3091 & 472 & 80.11 & 84.73 & 88.28 & 7 & 225 &  & 2.90 & 0.007 & 2.91 \\
\code{transportedIntegral\_co\ldots} & 2413 & 519 & 9926 & 2242 & 78.49 & 77.41 & 85.30 & 6 & 313 & \checkmark & 2.90 & 0.037 & 2.94 \\
\code{step1\_double} & 2315 & 189 & 12977 & 1586 & 91.84 & 87.78 & 93.21 & 8 & 260 &  & 2.94 & 0.015 & 2.95 \\
\code{holomorphic\_constant\_r\ldots} & 1756 & 558 & 51168 & 22596 & 68.22 & 55.84 & 74.69 & 6 & 280 & \checkmark & 2.88 & 0.064 & 2.94 \\
\code{exists\_chartMatchingCo\ldots} & 1645 & 313 & 11483 & 1938 & 80.97 & 83.12 & 88.03 & 5 & 233 & \checkmark & 2.86 & 0.023 & 2.88 \\
\code{insertZero\_nonadjacent\ldots} & 1509 & 299 & 6122 & 889 & 80.19 & 85.48 & 88.56 & 6 & 238 & \checkmark & 2.74 & 0.013 & 2.76 \\
\code{dfsNode\_dom} & 2879 & 1448 & 4191 & 1442 & 49.70 & 65.59 & 71.76 & 6 & 188 & \checkmark & 2.78 & 0.015 & 2.80 \\
\code{schemaTruthStage\_imp} & 746 & 80 & 194078 & 4396 & 89.28 & 97.73 & 95.67 & 7 & 240 &  & 2.91 & 0.018 & 2.93 \\
\code{cechCount} & 911 & 170 & 157384 & 6424 & 81.34 & 95.92 & 92.42 & 5 & 185 &  & 2.87 & 0.027 & 2.90 \\
\addlinespace[2pt]\multicolumn{14}{l}{\emph{PutnamBench (v4.26--4.27)}} \\
\code{putnam\_1964\_a4} & 1769 & 78 & 111469 & 2028 & 95.59 & 98.18 & 97.92 & 10 & 188 &  & 2.77 & 0.020 & 2.79 \\
\code{putnam\_1995\_a3} & 4085 & 116 & 14126 & 874 & 97.16 & 93.81 & 96.99 & 9 & 215 &  & 2.77 & 0.017 & 2.79 \\
\code{putnam\_1988\_a2} & 2162 & 96 & 58092 & 2429 & 95.56 & 95.82 & 97.13 & 16 & 208 &  & 2.82 & 0.033 & 2.85 \\
\code{putnam\_2001\_b1} & 3178 & 100 & 50140 & 1828 & 96.85 & 96.35 & 97.73 & 9 & 230 &  & 2.84 & 0.012 & 2.85 \\
\code{putnam\_1964\_b2} & 1754 & 27 & 134465 & 797 & 98.46 & 99.41 & 99.29 & 22 & 298 &  & 2.83 & 0.013 & 2.84 \\
\code{putnam\_2007\_a1} & 5464 & 75 & 528122 & 10697 & 98.63 & 97.97 & 98.87 & 7 & 265 &  & 2.72 & 0.033 & 2.75 \\
\code{putnam\_1970\_b5} & 2134 & 68 & 288942 & 4230 & 96.81 & 98.54 & 98.45 & 9 & 254 &  & 2.90 & 0.022 & 2.92 \\
\code{putnam\_1968\_a1} & 1084 & 76 & 60241 & 3451 & 92.99 & 94.27 & 95.75 & 14 & 193 &  & 2.68 & 0.054 & 2.73 \\
\code{putnam\_1991\_b1} & 3313 & 150 & 342800 & 6626 & 95.47 & 98.07 & 97.85 & 9 & 233 &  & 2.78 & 0.015 & 2.79 \\
\code{putnam\_2019\_a1} & 2381 & 88 & 3223935 & 11701 & 96.30 & 99.64 & 98.65 & 12 & 194 &  & 2.81 & 0.015 & 2.83 \\
\midrule
Total / mean & 85453 & 11356 & 7438069 & 582723 & 80.48 & 83.49 & 87.99 & 563 & 13429 & 17 & 168.30 & 1.59 & 169.89 \\
    \bottomrule
  \end{tabular}}
\end{table}

\begin{figure}[H]
\begin{Verbatim}[fontsize=\small,frame=single]
theorem putnam_1964_b2
(S : Type*) [Fintype S] [Nonempty S]
(P : Finset (Set S))
(hPP : ∀ T ∈ P, ∀ U ∈ P, T ∩ U ≠ ∅)
(hPS : ¬∃ T : Set S, T ∉ P ∧ (∀ U ∈ P, T ∩ U ≠ ∅))
: (P.card = 2 ^ (Fintype.card S - 1)) := by
  apply Nat.mul_left_cancel Nat.two_pos
  rewrite [mul_pow_sub_one Fintype.card_pos.ne', Fintype.card_set.symm,
    ← Set.Intersecting.is_max_iff_card_eq] <;>
    simp only [Set.Intersecting, Set.disjoint_iff_inter_eq_empty]
  any_goals assumption
  clear hPP
  grind
\end{Verbatim}
  \caption{\code{putnam\_1964\_b2} from the submitted run: 27 tokens and 797\,kHB, down from 1{,}754 tokens and 134{,}499\,kHB (score 99.29). The line break in the \code{rewrite} list was added for width.}
  \label{fig:example}
\end{figure}

\section{Infrastructure and optimizer details}
\label{app:infra}

\begin{figure}[H]
  \centering
  \includegraphics[width=\linewidth]{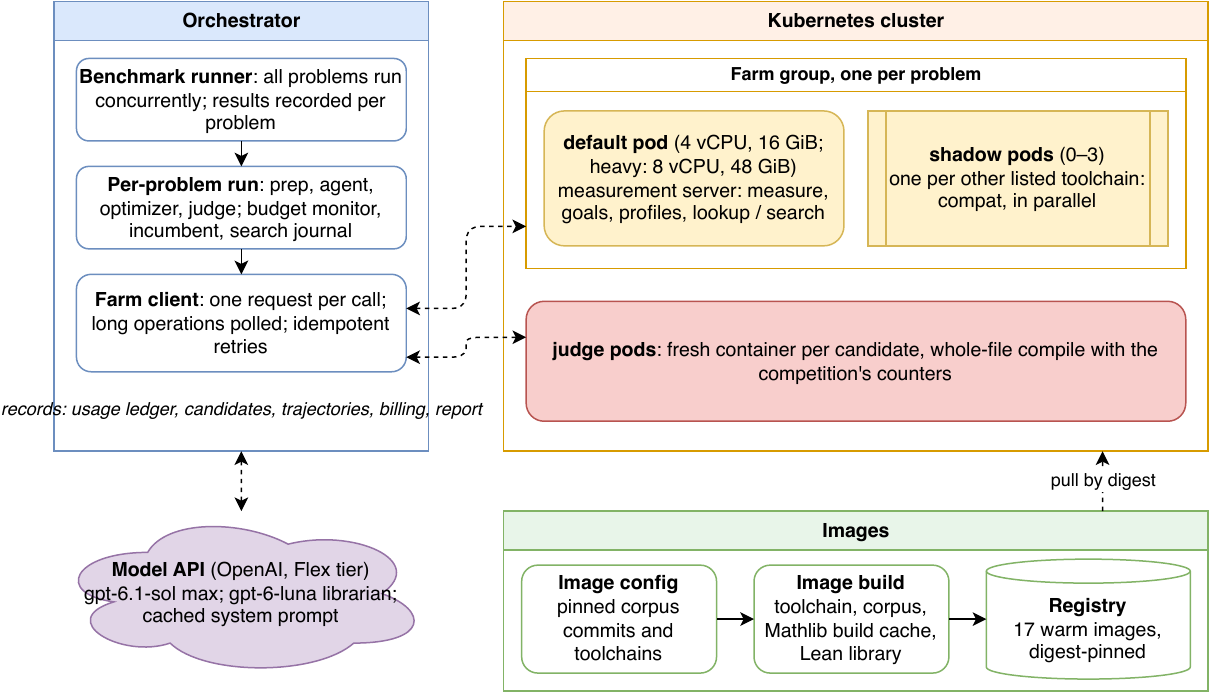}
  \caption{Infrastructure. One orchestrator runs all proofs concurrently. Each problem leases one warm pod per toolchain it lists (127 pods for the final benchmark), started from digest-pinned warm images, so every candidate is measured and compat-checked inline without compiling Mathlib.}
  \label{fig:infra}
\end{figure}

\begin{table}[H]
  \caption{The deterministic optimizer, no API spend. ``Seed'' is the proof it starts from. Gains are judge-score differences against the seed. Rows above the rule are warm-up runs (row numbers refer to Table~\ref{tab:ladder}); rows below it are final-benchmark runs (Table~\ref{tab:final-ladder}).}
  \label{tab:optimizer}
  \centering
  \footnotesize
  \begin{tabular}{llrr}
    \toprule
    Seed & Optimizer version & Seed score & Gain \\
    \midrule
    Original proofs & lint, ablate, minlemmas, trybotup & 33.33 & +10.35 \\
    agent output, first harness & same & 86.90 (farm) & +0.15 (farm) \\
    agent output, row 4 harness & + InfoTree phases (slim, share, inline) & 88.58 & +0.10 \\
    agent output, row 6 harness & same & 91.07 & +0.03 \\
    agent output, row 6 harness & + golf, clean & 91.07 & +0.21 \\
    row 6 incumbent at \$1 / \$2 & + golf, clean & 88.62 / 90.48 & +0.43 / +0.20 \\
    agent output, row 9 harness & + hinted collapse & 92.36 & $\le$ +0.05 \\
    agent output, row 9 harness & + term substitution & 92.36 & 0 (15/15 unchanged) \\
    \midrule
    run 1 agent output & golf, clean & 84.70 & +0.11 \\
    run 1 agent output & + structural passes & 84.70 & +0.12 \\
    run 4 agent output & same (submitted) & 87.92 (farm) & +0.07 \\
    \bottomrule
  \end{tabular}
\end{table}

\section{The Lean metaprogramming layer in detail}
\label{app:lean}

\begin{figure}[H]
  \centering
  \includegraphics[width=0.85\linewidth]{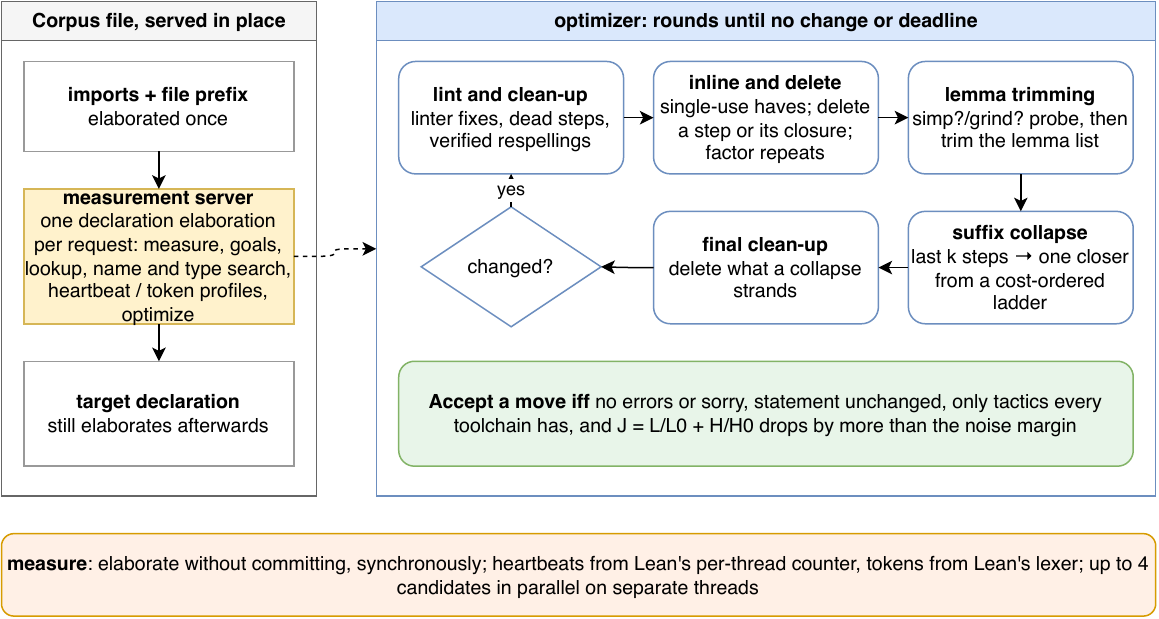}
  \caption{The Lean layer. The measurement server is spliced immediately before the target, so the file prefix is elaborated once and each request costs one declaration elaboration. Every agent tool except the librarian is a client of this server, and the optimizer (Stage~2) runs verified rewrites to a fixpoint inside it.}
  \label{fig:compressor}
\end{figure}

\paragraph{One elaboration per candidate.} The measurement server is a command elaborator spliced immediately before the target declaration. It reads newline-delimited JSON requests and answers each by elaborating a single declaration in the environment of the file prefix. After end of input, the original declaration still elaborates, so splicing the server in never breaks the file. The median evaluation, transport included, took 5.7\,s, against about 4.5\,min for a judge run (a whole-file compile in a fresh container, pod start included); that gap is what makes thousands of evaluations per run affordable.

\paragraph{Exact measurement.} The server elaborates a candidate synchronously (\code{Elab.async false}) inside \code{withoutModifyingEnv}, so nothing it declares survives. It also captures and restores the message log, so its diagnostics never leak into the file. Length is Lean's token function run in the target environment, so notation declared by imports lexes exactly as the judge lexes it. Heartbeats are a difference of \code{IO.getNumHeartbeats}, which reads the calling thread's counter. So the server can elaborate up to four candidates on dedicated threads without their counts interfering. Batched counts agree with serial measurement from the same snapshot to within a few raw heartbeats.

\paragraph{Reading the InfoTree.} For every tactic it runs, Lean records the goals before and after and the term produced. The library indexes this \code{InfoTree} by source span into one node per tactic occurrence. Each node records:
\begin{itemize}
  \item its goal counts;
  \item how many times it ran (once per goal under \code{<;>});
  \item whether it ever met an open goal;
  \item the complete term it produced for its first goal, which amounts to \code{show\_term} for every closer at once.
\end{itemize}
From this index the server derives three things. The first is goal snapshots for every branch (the agent's goal-state tool). The second is the steps that never ran, which the optimizer's pruning phase deletes. The third is token attribution, in which every counted token belongs to exactly one innermost span, so the exclusive counts sum to the scored length. Heartbeat attribution comes from Lean's own profiler (\code{trace.profiler.useHeartbeats}). Tracing inflates costs by 1.3--1.4$\times$, so its numbers are diagnostics, never scores. Info trees themselves cost heartbeats, so indexing runs in a separate, unmeasured elaboration, and the server discards them after each request.

\paragraph{Retrieval in scope.} Name lookup, paginated name search and type-shape search run over the server's environment. By construction, that environment contains exactly the declarations visible at the theorem, so the agent cannot cite a lemma that appears later in the file. The type-shape index records the constants in each declaration's type and the head of its conclusion. A query ranks declarations by overlap with a goal's symbols and by a matching conclusion head. The index is built once, outside any measured elaboration.

\paragraph{The optimizer} is a deterministic search over verified edits that minimizes $J = L/\Lzero + H/\Hzero$, compared exactly in integers. A move is accepted only if the result meets four conditions:
\begin{itemize}
  \item it elaborates without \code{sorry};
  \item it leaves the statement unchanged;
  \item it uses only tactics on an allow-list of those that parse on every listed toolchain;
  \item it lowers $J$ by more than a noise margin.
\end{itemize}
Phases run cheapest first, each one shrinking what the expensive ones must search:
\begin{enumerate}
  \item linter fixes and dead-step pruning;
  \item verified respellings (argument holes, deleted named arguments and ascriptions, adjacent-tactic merges, positional constructor idioms);
  \item merging two \code{case} arms with identical bodies into one \code{case a | b => body}, and merging identical trailing branch arms into one \code{all\_goals};
  \item inlining single-use \code{have}s;
  \item deleting a step or its dependency closure, delta-debugging style;
  \item factoring repeated tactic runs (after inlining, since one removes names and the other introduces them);
  \item trimming \code{simp}/\code{grind} lemma lists to those \code{simp?}/\code{grind?} report, and dropping elements of any bracketed list;
  \item converting tactic blocks to terms;
  \item \textbf{suffix collapse}, which replaces the last $k \le 8$ steps of a block with one probe of a closer ladder;
  \item targeted node-level search at whichever steps the heartbeat profiler marks hottest, trying cheaper tactics in their place;
  \item replacing a closer with \code{apply}/\code{rw}/\code{simp only} of the lemma its term uses;
  \item a final deletion pass for the bindings a collapse strands.
\end{enumerate}
The closer ladder is a cheap \code{try?}: a cost-ordered sequence of closers (\code{rfl}, \code{omega}, \code{simp}, \code{grind}, \ldots). Its Mathlib-only rungs fail to parse on Mathlib-free corpora and are skipped there. Its \code{fun\_induction} rungs read the function to recurse on from the goal. Rounds repeat until nothing changes or a deadline passes, and the search returns the best proof found so far. Batched search accepts exactly what sequential search would. Appendix~\ref{app:trace} shows a trace on a benchmark proof.

\paragraph{Portability.} The library compiles on Lean v4.25 through v4.35-rc3. The \code{String} API changes across that range, so all text handling goes through \code{List Char} and our own trimming. A generated \emph{module} variant serves module files (v4.27 and later) whose proofs need \code{import all}. The same transforms are also exposed as editor commands (\code{\#optimize}, \code{\#profile}, \code{\#ablate}, \ldots), so a library maintainer can compress a proof in place without the agent.

\section{Development on the warm-up set and ablations}
\label{app:dev}

Table~\ref{tab:ladder} lists the warm-up runs at which the score changed by more than noise, one change per row. Two rows change the measurement rather than the pipeline: the bullet rule (row 6) and the pipe rule (row 9) made our counts match the arena's. Table~\ref{tab:ablations} lists the changes that did not help. On six hard warm-up problems, gpt-6-sol at low and medium effort stopped early with most of the budget unspent (55.1, 65.7); at high effort it reached 75.9, against 83.2 for gpt-6-astra.

\begin{table}[H]
  \caption{Development trajectory on the warm-up set; each row's change is relative to the row above. $^\dagger$Metered without cache-write pricing; billed cost was about 6\% higher and some proofs exceeded \$3. $^\ddagger$Measurement change. From row 4 on, every proof is within \$3. Row 10 is the agent configuration we submitted; the submission adds the deterministic optimizer.}
  \label{tab:ladder}
  \centering
  \footnotesize
  \setlength{\tabcolsep}{4pt}
  \begin{tabular}{llrrrr}
    \toprule
    Run & Change & Comb. & Len.\,\% & HB\,\% & Total \$ \\
    \midrule
    1 & gpt-6-astra (high); first harness; flat \$2.50 finalize; optimizer & 86.99 & 76.7 & 84.3 & 39.79$^\dagger$ \\
    2 & proof-body interface, located errors, librarian, break-even caps & 87.74 & 77.8 & 85.4 & 40.37$^\dagger$ \\
    3 & adaptive finalize; effort xhigh & 88.17 & 79.8 & 84.7 & 38.57$^\dagger$ \\
    4 & cache-write metering; tool fixes & 88.58 & 80.0 & 85.8 & 39.68 \\
    5 & \textbf{gpt-6.1-sol (max)} & 90.57 & 83.4 & 88.3 & 37.89 \\
    6 & bullet token rule$^\ddagger$; refreshed images & 91.07 & 84.4 & 88.9 & 35.89 \\
    7 & context-flush continuations; judge fix; optimizer & 91.12 & 84.0 & 89.4 & 41.26 \\
    8 & candidate archive, patch evaluation, direct lookup, type search; no optimizer & 91.97 & 86.2 & 89.7 & 42.65 \\
    9 & \textbf{Flex tier} (half price); pipe token rule$^\ddagger$ & 92.36 & 86.9 & 90.2 & 39.17 \\
    \textbf{10} & \textbf{warm-up-specific skill content removed} & \textbf{92.31} & \textbf{86.5} & \textbf{90.5} & \textbf{36.47} \\
    \bottomrule
  \end{tabular}
\end{table}

\begin{table}[H]
  \caption{Changes that did not improve the score. ``Base'' is the comparable run without the change. (6), (5), (12) and (16) mark rows evaluated on that many problems; the last block was run on final-benchmark problems.}
  \label{tab:ablations}
  \centering
  \footnotesize
  \setlength{\tabcolsep}{4pt}
  \begin{tabular}{lrrl}
    \toprule
    Change & Base & With & Reading \\
    \midrule
    \multicolumn{4}{l}{\emph{More information for the agent (warm-up)}} \\
    Lemma suggestions in \code{goals}; persistent librarian & 88.01/88.17 & 88.06 & noise \\
    Offline Loogle-style pattern search & 88.58 & 88.26 & noise \\
    Larger prompts and tool outputs & 88.20 & 87.56/87.07 & $-0.6$ to $-1.1$ \\
    Heartbeat profiles on every new best (12) & 91.94/92.03 & 91.98 & null \\
    \midrule
    \multicolumn{4}{l}{\emph{More structure in the search (warm-up)}} \\
    Forced-diversity phases (rethink, block, golf) & 91.28 & 90.08 & $-1.2$ \\
    Planner (astra) delegating to worker (sol) (6) & 83.21 & 80.76 & $-2.5$ \\
    Two \$1.50 attempts instead of one \$3 & 88.27 & 86.42 & $-1.9$ \\
    Restart from original after 3 stalls (5) & 88.00 & 88.34 & noise \\
    \midrule
    \multicolumn{4}{l}{\emph{Reasoning effort and prompt content (warm-up)}} \\
    gpt-6.1-sol effort xhigh instead of max & 92.31 & 92.24 & noise, $-3\%$ cost \\
    Remove golf-pattern catalogue & 92.36 & 92.31 & noise \\
    \midrule
    \multicolumn{4}{l}{\emph{Final-benchmark pilots}} \\
    Compiler-verified research, four helper models (6) & 76.86 & 75.2--75.8 & $-1.0$ to $-1.7$ \\
    Agent model as librarian (6) & 76.86 & 74.32 & $-2.5$ \\
    Skill with 30 extra golf idioms (16) & --- & +0.14 & noise ($\pm 0.30$) \\
    \bottomrule
  \end{tabular}
\end{table}

\section{Agent tool interface}
\label{app:tools}

Tool results are JSON with a size limit per field, so a truncated result still parses and says what was omitted and how to retrieve it. Once the finalize rule fires, every result carries a notice to finish. Table~\ref{tab:tools} lists the tools and how often the agent called them in the submitted run.

\begin{table}[H]
  \caption{Agent tools and call counts in the submitted run (10{,}679 agent turns over 60 problems).}
  \label{tab:tools}
  \centering
  \small
  \begin{tabular}{lp{0.5\linewidth}r}
    \toprule
    Tool & Purpose & Calls \\
    \midrule
    Patch evaluation & exact replacements on an archived candidate, batched variants & 3{,}396 \\
    Name search & paginated name search with namespace and kind filters & 2{,}584 \\
    Name lookup & signatures and sources of named constants & 2{,}467 \\
    Proof evaluation & compile and measure 1--4 whole proof bodies & 2{,}366 \\
    Candidate archive & archive listing and retrieval by ID & 2{,}144 \\
    Goal states & goal state per tactic and branch; paged retrieval & 1{,}291 \\
    Notes & persistent notes (8\,KiB global, 2\,KiB per branch) & 1{,}008 \\
    Full-text lookup & full text of a long signature or source & 756 \\
    Submit & end the session & 419 \\
    Librarian & question answered by gpt-6-luna with the lookup tools & 293 \\
    Heartbeat profile & heartbeats per tactic (Lean's heartbeat profiler) & 288 \\
    Type-shape search & type-shape retrieval over the environment & 161 \\
    Token profile & tokens per source span, exclusive counts sum to $L$ & 97 \\
    \bottomrule
  \end{tabular}
\end{table}

The skill document in the system prompt is part of the code archive. It contains no budget information and no benchmark proofs.

\section{The optimizer on a benchmark proof}
\label{app:trace}

We ran the editor command \code{\#optimize (tokens := 313) (heartbeats := 5175) in} on the original proof of Strata's \code{substOldPostSubset} (a warm-up problem), inside the problem's Lean v4.26 warm image. Stage~2 runs the same search, but under a deadline and a cross-toolchain tactic allow-list; the editor command uses neither. This trace predates the branch-merging, term-conversion and closer-replacement phases.

The search reached a fixpoint after three rounds: 313 $\to$ 189 tokens and 5{,}175 $\to$ 3{,}188\,kHB, from 42 accepted moves out of 2{,}302 verified candidates. That scores about 59, against the original's 33.33. The agent's proof of the same theorem in a warm-up run has 34 tokens and 402\,kHB (score 93.79). It is a different argument, which no sequence of local edits reaches.

The trace below is the command's output for the first round, followed by every later phase that made a move. Each line shows the moves accepted out of the candidates elaborated, and the cost after the phase. In the phase names, \code{prune} deletes dead steps, \code{golf}/\code{clean}/\code{slim} are verified respellings, \code{ablate} and \code{share} delete and factor steps, \code{minlemmas} trims lemma lists, \code{whole} tries a single closer for the whole proof, and \code{trybotup} is the suffix collapse.

\begin{Verbatim}[fontsize=\footnotesize,frame=single]
#optimize (3 round(s)): 313 -> 189 tokens (39%), 5179 -> 3188 kHB (38%)
                        [42 move(s) from 2302 candidate(s)]
round 1
  lint:      0 move(s)/0 cand    -> 313 tok 5175 kHB  from 0 hint(s)
  prune:     9 move(s)/1 cand    -> 295 tok 5171 kHB  from 9 unreached
  golf:      0 move(s)/3 cand    -> 295 tok 5171 kHB
  clean:     0 move(s)/5 cand    -> 295 tok 5171 kHB
  slim:      2 move(s)/19 cand   -> 291 tok 5172 kHB
  inline:    0 move(s)/0 cand    -> 291 tok 5172 kHB
  ablate:   15 move(s)/191 cand  -> 224 tok 5037 kHB
  share:     0 move(s)/24 cand   -> 224 tok 5037 kHB
  minlemmas: 4 move(s)/1038 cand -> 232 tok 3913 kHB
  whole:     0 move(s)/3 cand    -> 232 tok 3913 kHB
  trybotup: 10 move(s)/444 cand  -> 194 tok 3187 kHB
  ablate2:   0 move(s)/75 cand   -> 194 tok 3187 kHB
round 2
  slim:      2 move(s)/15 cand   -> 189 tok 3188 kHB
round 3
  (no phase moves; fixpoint)
\end{Verbatim}

The phases divide the work as designed:
\begin{itemize}
  \item \code{prune} deletes, in a single candidate, the nine tactic steps that the \code{InfoTree} shows never met a goal.
  \item \code{ablate} deletes 15 more steps that ran but were unnecessary.
  \item \code{minlemmas} replaces \code{simp} calls with the lemma lists \code{simp?} reports. This trades 8 tokens for 1{,}124\,kHB, a trade worth making at this problem's exchange rate ($\Hzero/\Lzero \approx 16.5$\,kHB per token).
  \item \code{trybotup} collapses ten trailing blocks into single closers.
\end{itemize}
The first five cases of the induction show the effect:

\begin{Verbatim}[fontsize=\footnotesize,frame=single]
-- original (excerpt)                       -- after #optimize (excerpt)
induction post <;> simp [substOld]         induction post <;> simp only [substOld]
case fvar =>                               case fvar =>
  intros x Hin                               intros x Hin
  rename_i m name ty2                        simp_all
  simp at m                                · solve_by_elim
  simp at name                             · solve_by_elim
  simp at ty2                              · solve_by_elim
  simp_all                                 · assumption
case op =>
  intros x Hin
  rename_i m name ty2
  simp at m
  simp at name
  simp at ty2
  simp_all
case const =>
  intros x Hin
  rename_i m name
  simp at m
  simp_all
case bvar =>
  intros x Hin
  rename_i m d
  simp at m
  simp_all
case abs ih =>
  exact ih
\end{Verbatim}


\end{document}